# Bayesian Convolutional Neural Networks for Seven Basic Facial Expression Classifications

Yuan Tai    Yihua Tan    Wei Gong    Hailan Huang

The seven basic facial expression classifications are a basic way to express complex human emotions and are an important part of artificial intelligence research. Based on the traditional Bayesian neural network framework, the ResNet18_BNN network constructed in this paper has been improved in the following three aspects: (1) A new objective function is proposed, which is composed of the KL loss of uncertain parameters and the intersection of specific parameters. Entropy loss composition. (2) Aiming at a special objective function, a training scheme for alternately updating these two parameters is proposed. (3) Only model the parameters of the last convolution group. Through testing on the FER2013 test set, we achieved 71.5% and 73.1% accuracy in PublicTestSet and PrivateTestSet, respectively. Compared with traditional Bayesian neural networks, our method brings the highest classification accuracy gain.

1. Introduction

With the development of society and the progress of science and technology, images, as a rich visual signal, play an increasingly important role in the field of scientific research, and facial expression images are undoubtedly a research hotspot among them. As early as the 1970s, psychologists led by Ekman [2] divided the basic human emotions, including: Anger, Disgust, Fear, Happiness, Sadness, Surprise. In order to make the expression classification more perfect, in subsequent research, people added new categories such as neutral. Because facial expressions are an intuitive manifestation of emotions, these seven categories are widely accepted by scholars who study facial expression recognition.

After discretizing expressions into several categories, expression recognition can be regarded as a common classification problem. In addition to classic classifiers such as SVM and random forest in machine learning, artificial neural networks can handle classification problems in an end-to-end manner. For neural networks, the model parameters of ordinary neural networks are fixed, so the model itself does not have due uncertainty. With the continuous deepening of research, in order to make the neural network compatible with more uncertainties, the Bayesian Neural Network (BNN) is further proposed on this basis.

Bayesian-CNNs embed probability distributions into the model to provide uncertainty representation against overfitting, which can be essentially seen as a regularization of parameters. It's often to optimize the model by inferring the posterior probabilities of the parameters of the network. The posterior inference in a Bayesian-CNNs is often implemented by minimizing the Kullback-Leibler divergence between the approximate distribution and the true posterior. [1] In fact, Bayesian-CNNs modeling the epistemic uncertainty on all the parameters in each convolutional layer brings an optimization difficulty that the parameters may be far from the optimal point. Experimental analysis on VGG19[5] based Bayesian-CNNs indicate that the closer to the output layer, the smaller the standard deviation of parameters of a trained Bayesian-CNNs is. Therefore, we speculate that the reason why the Bayesian-CNNs have limited performance improvement in classification is that the parameters of the layers close to the input are not well-trained because of the uncertainty modeling.

Based on the above analysis, this paper models the cognitive uncertainty of different convolutional groups and compares the results between them, and uses uncertainty to model the convolutional layer of a specific group to achieve the highest accuracy. A new objective function is proposed, which is composed of the KL loss of uncertain parameters and the cross-entropy loss of some parameters. In order to process these two parameters at the same time, a training scheme that alternately updates them is used.

The organization of this paper is as follows: In section 2 shows the background of our method. Section 3 indicates the details of our method. Section 4 shows our experimental results on FER2013[12] database, and Section 5 states the conclusion.

2. Background
2.1. Bayesian convolutional neural networks

In the deterministic CNNs, we assume a training set $D = (x, y)$. Using $D$ we are interested in learning an inference function $f_w$ with parameters w and the estimated labeld $\hat{y}$ is obtained as follows:

$$\hat{y} = \arg\max_y f_w(x, y) \tag{1}$$

Differently, Bayesian-CNNs are probabilistic models that introduce a probability distribution on the parameters to model the epistemic uncertainty. In Bayesian modeling, in contrast to finding a point estimate of the model parameters, the idea is to estimate an (approximate) posterior distribution of the parameters $p(w|D)$ to be used for probabilistic prediction:

$$p(y|x, D) = \int f_w(x, y) p(w|D) dw \tag{2}$$

where $f_w(x, y) = p(y|x, w)$. The predicted label $\hat{y}$ can be accordingly obtained by sampling $p(y|x, D)$ or taking its maxima.

## 2.2. Variational Approximation

To obtain the posterior distribution of the parameters, a common approach is to learn a parameterized approximating distribution $q(w|\theta)$ that minimizes Kullback-Leibler(KL) divergence with the true Bayesian posterior on the parameters which can be express as:

$$\theta^* = \arg\min_\theta \text{KL}(q(w|\theta) \parallel P(w|\mathcal{D})) \tag{3}$$

This objective function can be written as:

$$\mathcal{L}_{\text{unc}}(\theta, D) \approx \text{KL}[q(w|\theta) \parallel P(w)] - \mathbb{E}_{q(w|\theta)}[\log(P(\mathcal{D}|w))] \tag{4}$$

Further, the below equation can be approximated using N Monte Carlo samples $w_1$ from the variational posterior [1].

$$\mathcal{L}_{\text{unc}}(\theta, D) \approx \sum \log q(w^i|\theta) - \log P(w^i) - \log(P(\mathcal{D}|w^i)) \tag{5}$$

In this work, this approximation method is used to learn the uncertain parameters.

## 3. The Proposed Method
### 3.1. Structure of Networks

To study the question that where to model epistemic uncertainty, we only conduct uncertainty modeling for each network in a specific convolution group. The reasons are as follows: 1) since we aim to explore the influence of location on the uncertainty modeling, we need to eliminate the interference between different convolution groups; 2) we hope to minimize the increase of parameters to reduce the difficulty of training. In this paper, the Gaussian distribution is placed over parameters of different convolution groups to model epistemic uncertainty. Fig.1 shows the structure that modeling epistemic uncertainty over the parameters of the last convolution group of ResNet18. The first four convolution group represented by yellow blocks is the certain part whose parameters are a value expressed by $w_1$. The last convolution group represented by the blue block is the uncertain part whose parameters are the probability distribution expressed by $w_2$. Besides, we replace the original fully connected(FC) layers with one FC layer to reduce the number of parameters. Further, the Batch Normalization (BN)[3] layer is added after each convolution layer which is activated by ReLU [4] for increasing the convergence speed.

### 3.2. Loss function

In our model, the parameters consist of $w_1$ and $w_2$ for the certain part and the uncertain part respectively so that the likelihood function(Refer to section 2.2) should be the joint probability distribution of $w_1$ and $w_2$. Besides, $w_2$ is obtained by sampling the distribution $\theta$. Therefore, the objective function becomes to the following form:

$$\mathcal{L}_{\text{unc}}(\theta, D, w_1) = \text{KL}[q(w_2|\theta) \parallel P(w_2)] - \mathbb{E}_{q(w_1, w_2|\theta)}[\log(P(\mathcal{D}|w_1, w_2))] \tag{6}$$

Thereby, the Monte Carlo samples from the variational posterior becomes:

$$\mathcal{L}_{\text{unc}}(\theta, D, w_1) \approx \sum \log q(w_2^i|\theta) - \log P(w_2^i) - \log(P(\mathcal{D}|w_1, w_2^i)) \tag{7}$$

Finaly, the total objective function can be expressed as:

$$J(\theta, D, w_1) = L_{cen} + L_{unc} \tag{8}$$

$L_{cen}$ is the cross-entropy loss in this paper which is used to update the certain parameters $w_1$.

$L_{cen}$ uses the common softmax loss function based on multi-classification. Suppose the input is an emoticon image $I$, where $y^*$ is the image label, and $y$ is the predicted value of the network output, which is generally represented by one-hot encoding. The loss function formula $L_{cen}$ is shown in Eq.9.

$$L_{cen} = -\sum_{j=1}^{N} y_j^* \log y_j \qquad (9)$$

In the classification loss $L_{cen}$, $N$ represents the total number of all categories to be detected. Fig. 2. shows the inputs of two kinds of loss function.

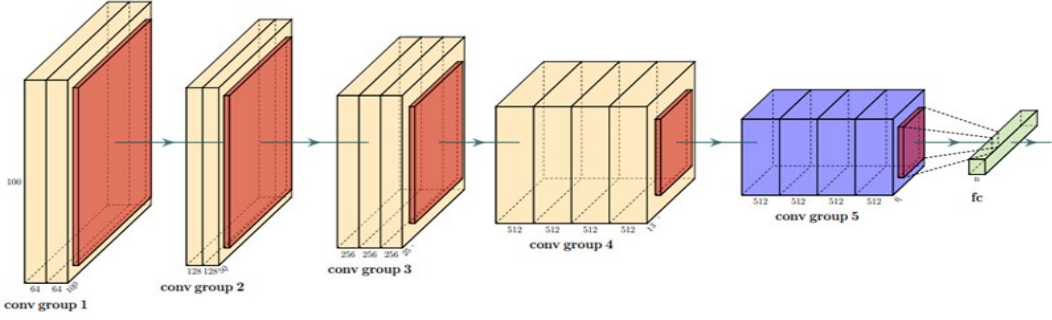

Figure 1. The network structure that modeling epistemic uncertainty over the parameters of the last convolution group of ResNet18.Compared to ResNet18, the network poses several differences: (1) the last three Fully Connected (FC) layers were replaced by one FCwhose dimension is equal to the number of classes; (2) the last convolution group consists of four convolution layers with uncertainparameters. (3) the Batch Normalization (BN)[3] layer is added after each convolution layer which is activated by ReLU [4].

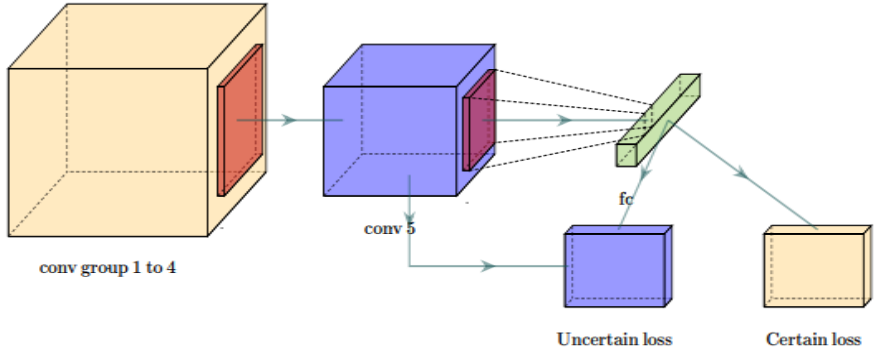

Figure 2. Conv 5 is the convolutional group modeled by epistemicuncertainty. The block of certain loss represents the first term in Eq. 8, and the block of uncertain loss represents the second term.

### 3.3. Training strategy

Blundell [1] propose an effective approximating method to obtain the posterior distribution of uncertain parameters of Bayesian-CNNs which are also used in this paper. Differently, since our model contains both certain parameters and uncertain parameters which are needed to be updated in different ways, we update them alternately. Specifically, the training scheme contains two parts, the uncertain part, and the certain part. Firstly, uncertain parameters are updated by the following steps:

**Forward propagation stage:**

1. For uncertain convolutional layer, sampling is carried out:

$$w_2 = \mu + \log(1 + \exp(\rho)) \circ \epsilon, \epsilon \sim N(0, I) \qquad (10)$$

where the uncertain parameters $\theta$ of the network is parametrized by $\theta = (\mu, \rho)$. Standard deviation is parametrized as $\sigma = \log(1 + \exp(\rho))$ to ensure the positive value. $\epsilon$ is the noise drawn from unit Gaussian and $\circ$ is a pointwise multipliation.

2. Calculate the uncertain loss $L_{unc}$.
**back propagation stage:**
1. As for uncertain layers, calculate the gradient of the mean and standard deviation:

$$\Delta_\mu = \frac{\partial L_{unc}(\theta, D, w_1)}{\partial w_2} + \frac{\partial L_{unc}(\theta, D, w_1)}{\partial \mu} \quad (11)$$

$$\Delta_\rho = \frac{\partial L_{unc}(\theta, D, w_1)}{\partial w_2} \frac{\epsilon}{1 + \exp(-\rho)} + \frac{\partial L_{unc}(\theta, D, w_1)}{\partial \rho}$$

2. Update the parameters ($\eta$ means learning rate):

$$\mu_{new} = \mu_{old} - \eta \Delta_\mu, \rho_{new} = \rho_{old} - \eta \Delta_\rho \quad (12)$$

Secondly, certain parameters are updated by the backpropagation method. At this time, the value of the uncertain parameters is fixed as the mean value. The total training scheme is outlined in Algorithm 1.

---
**Algorithm 1 Learning of $w_1$ and $w_2$**

1 Input: Dataset $D_{train} = (x_i, y_i)$, learning rate $\alpha$
2 for i=1 to N do:
3     Samping $w_2$ with Eq. 10
4     Calculate log $\log q(w_2^i|\theta) - \log P(w_2^i) - \log(P(\mathcal{D}|w_1, w_2^i))$
5 Calculate $L_{unc}$ with Eq. 7.
6 Calculate the gradient of $\mu$ and $\rho$ with Eq. 11
7 Update $\mu$ and $\rho$ to minimize $L_{unc}$ with Eq. 12.
8 Fix the value of $w_2$ as $\mu$
9 Calculate $L_{cen}$ with cross entropy.
10 Update $w_1$ to minimize $L_{cen}$ with backpropagation(BP).

---

## 4. Experiments
### 4.1. The impact of uncertainty introduced by different network layers

When constructing the network, this paper chooses to add uncertainty in part of the network layer. In order to verify the rationality of the design of ResNet18_BNN, the following experiments are done here for the location factor of uncertainty. In Fig. 3, five groups of different network configurations are listed, and their respective training curves are given. At the same time, the PublicTestSet of FER2013 is selected as the verification set to indicate the test accuracy of different networks. It can be seen that the training curve of group e converges faster and has an advantage in recognition rate, which fully verifies the rationality of this scheme.

### 4.2. Performance Testing

In experiments we choose ResNet18 for feature extraction of images, and form a lightweight and effective ResNet18_BNN network under the framework of Bayesian neural network.

In order to verify the effectiveness of the ResNet18_BNN network, we conducted experiments on the large field dataset FER2013.S. we compare this network method with other methods to illustrate the effectiveness of this method.

In Table 1, the comparison results of the ResNet18_BNN network and other network methods are listed. Among them, the single network method includes: CNN method (CNN+SVM) combined with SVM classifier [7]; Deep learning method based on correlation learning (DNNRL) [6]. The multi-task network method uses the Siamese nets structure (Siamese nets) [8], which combines the technology of face registration and feature fusion, and trains on three heterogeneous label data sets, which contain "gender" and "posture"," "emoji" and "age" these four attributes. Integrated network methods include: synthesizing 36 methods (36 CNNs) with different input depths [9]; combining alignment mapping networks with discriminative deep convolutional neural networks (AMNs+DNNs) [10]; and comprehensive considerations VGG, Inception, and ResNet network structures, and integrated tuning methods for multiple groups of models (Multiple models) [11].

It can be seen that the effect of the complex network is better than that of the ordinary network, and the ResNet18_BNN in this article only makes minor changes to the single network, but the accuracy obtained is more competitive. In addition, in the top three performance methods (Multiple models, Siamese nets, 36CNNs), not only the network structure is complex, but also auxiliary data sets and auxiliary networks are used. In contrast, ResNet18_BNN only performs experiments under given conditions , Get more robust and reliable results, so it has the characteristics of light and effective. At the same time, there are reasons to believe that by choosing a better-performing basic network, coupled with uncertainty modeling, the experimental results obtained will be further improved.

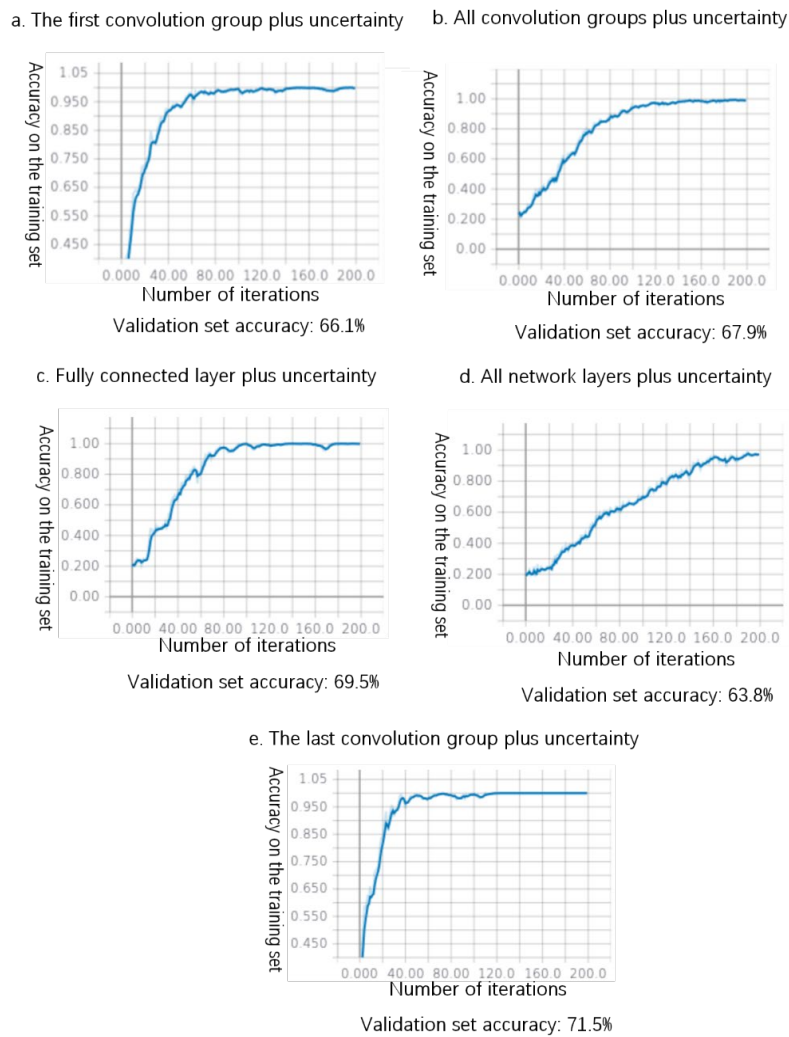

Figure 3. Experimental results of different uncertain networks (using PublicTestSet as the verification set)

Table 1 Experimental results of different network methods on the FER2013 data set

|  |  | Single network |  | Multitasking network | Integrated network |  |  | Resnet18_BNN |
|---|---|---|---|---|---|---|---|---|
| Algorithm |  | CNN+SVM | DNNRL | Siamese nets | 36 CNNs | AMNs+DCNs | Multiple models |  |
| Precision (%) | PublicTestSet | 69.4 | None | None | None |  |  | 71.5 |
|  | PrivateTestSet | 71.2 | 71.3 | 75.1 | 72.7 | 73.7 | 75.2 | 73.1 |

## 5. Conclusion

In this paper, we propose a ResNet18_BNN to classify human facial expressions into seven basic categories. We proposed a new objective function, which is composed of the KL loss of uncertain parameters and the cross-entropy loss of some parameters. Besides, aiming at a special objective function, we proposed a training scheme for alternately updating these two parameters. In order to achieve the best accuracy, ResNet18_BNN only model the parameters of the last convolution group. Through testing on the FER2013 test set, our method brings the highest classification accuracy gain. so it's an effective alternative to solve facial expression classification problem.